\documentclass[11pt, a4paper, logo, copyright, nonumbering]{xiaomi}

\usepackage[authoryear, sort&compress, round]{natbib}
\usepackage{dblfloatfix}
\usepackage{ulem}
\usepackage{caption}
\usepackage{dramatist}
\usepackage{xspace}
\usepackage{pifont}
\usepackage{multirow}
\usepackage{tcolorbox}
\usepackage{xltabular}
\usepackage{longtable}
\usepackage{hyperref}
\usepackage{wrapfig}
\usepackage{graphicx}

\usepackage{amsfonts}
\usepackage{amsmath}
\usepackage{amssymb}
\usepackage{lineno}
\usepackage{multirow}
\usepackage{adjustbox}

\usepackage[bottom]{footmisc}
\usepackage{algorithm}
\usepackage{algpseudocode}
\usepackage{CJKutf8}
\usepackage{subcaption}
\usepackage{setspace}
\usepackage{makecell}
\usepackage{graphicx}
\usepackage{multicol}
\usepackage{xspace}

\usepackage{cleveref}

\algnewcommand{\Initialize}{\textbf{Initialize: }}

\defcitealias{nondeter}{T. M. Lab}

\definecolor{xiaomiorange}{HTML}{FF6901}

\begin{abstract}
    Long-horizon and multi-turn agents typically generate short actions and process long observations from tools and environments. 
This growing context demands \textit{efficient prefill}, \textit{compact KV-cache storage}, and \textit{accurate long-context retrieval}.
To meet these demands, we introduce \textbf{HySparse2}, a hybrid sparse attention architecture with \textbf{two-level KV sharing}.
At the outer level, \textbf{KV Bridging} adopts a YOCO-style self-decoder and cross-decoder structure, but bridges only full-attention layers.
The self-decoder uses hybrid sliding-window attention (SWA), while the cross-decoder uses hybrid sparse attention.
The KV caches for full-attention layers in the cross-decoder are generated from the hidden states of full-attention layers in the self-decoder.
At the inner level, HySparse2 retains HySparse's core \textbf{KV Reuse} design with two refinements.
First, it replaces block-level sparsity with \textit{token-level} sparsity for finer long-context retrieval.
Second, it removes the separate SWA branch from sparse layers and instead forces a sliding window of recent tokens into the sparse selection.
This two-level KV sharing allows all cross-decoder KV caches to be constructed from self-decoder hidden states.
Prefill can therefore exit after the self-decoder, skipping all cross-decoder layers.
On an 80B-A3B MoE model, HySparse2 outperforms HySparse and Hybrid SWA on long-context retrieval and multi-turn agentic tasks, while substantially reducing prefill computation and KV-cache storage.

\end{abstract}

\begin{document}

{
\renewcommand{\absfont}{\linespread{1.08}\fontsize{11}{12}\selectfont}
\bgroup
\setlength{\parindent}{0pt}
\vspace*{-8pt}
\begin{adjustwidth}{0pt}{0pt}
\begin{center}
{\titlefont HySparse2: Hybrid Sparse Attention \\ with Two-Level KV Sharing \par}
{
\vskip3pt
{\normalfont\sffamily\fontsize{11}{15}\selectfont
Jianyu Wei$^{*}$ ~~~
Yizhao Gao$^{*}$ ~~~  %
Qihao Zhang ~~~
Shimao Chen ~~~
Zhengju Tang \\
Yu Cheng ~~~
Shengjie Zhou ~~~
Zihan Jiang ~~~
{\normalfont\sffamily\fontsize{11}{15}\selectfont
Yifan Song ~~~
Hailin Zhang \\
Liang Zhao ~~~
Bo Yang ~~~
Gang Wang ~~~
Shijie Cao$^{\diamond}$ ~~~
Fuli Luo$^{\diamond,\dagger}$} \\
\vskip7pt
{\normalfont\sffamily\fontsize{11}{15}\selectfont LLM-Core Xiaomi}
\vskip7pt
}
}
\end{center}
\end{adjustwidth}
\egroup
{\abscontent}
\thispagestyle{firststyle}
}

\renewcommand{\thefootnote}{\fnsymbol{footnote}}
\footnotetext[0]{$^{*}$Equal contribution.\quad $^{\diamond}$Corresponding authors.\quad $^{\dagger}$Team Lead.}
\renewcommand{\thefootnote}{\arabic{footnote}}

% Keep this nonfloating so it stays between the abstract and introduction.
\par\addvspace{8pt}
\noindent\begin{minipage}{\linewidth}
    \centering
    \hypersetup{hidelinks}
    \includegraphics[width=\linewidth]{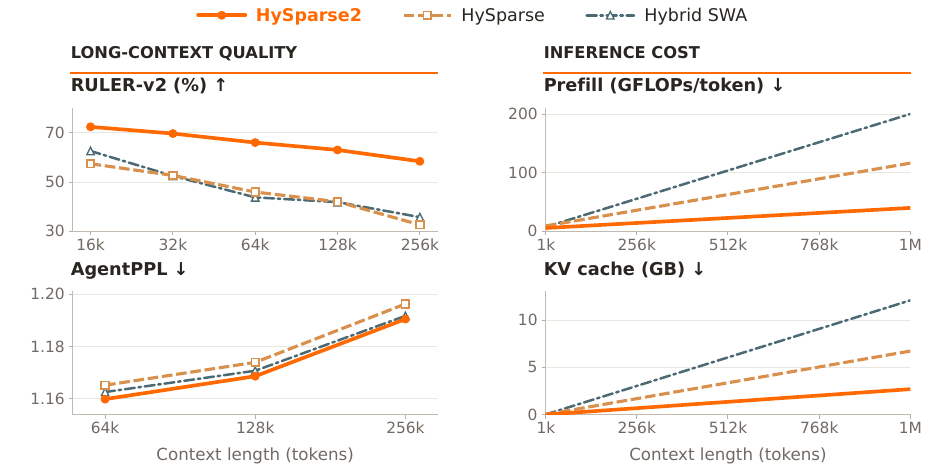}
    \captionsetup{font={footnotesize,stretch=1},skip=4pt,justification=centering,singlelinecheck=true,hypcap=false}
    \captionof{figure}{\textbf{HySparse2 delivers better long-context performance at lower prefill cost and smaller KV-cache storage.}}
    \label{fig:overview}
\end{minipage}\par
\newpage

\section{Introduction}
Agentic inference combines long contexts with multi-turn interaction.
Across interaction rounds, a short generated action or tool call can return a much longer search result, execution trace, or document that requires prefill before decoding resumes.
As observations accumulate, agents must retrieve and combine evidence across an expanding history, increasing both attention computation and KV-cache storage.
These workloads therefore demand efficient prefill, compact KV-cache storage, and accurate long-context retrieval.

HySparse~\citep{hysparse} addresses long-context efficiency by interleaving full-attention layers with sparse-attention layers.
Each full-attention layer supplies both selection indices and a KV cache to the following sparse layers, reducing attention computation and KV-cache storage without distilling an auxiliary indexer module~\citetext{\citealp{seerattn_v1}; \citealp{dsa}}.
For agentic inference, however, HySparse still leaves substantial room to shorten prefill, further reduce KV-cache storage, and improve retrieval precision.

In this work, we introduce \textbf{HySparse2}, which extends HySparse with \textbf{two-level KV sharing}.
Following YOCO~\citep{yoco}, the backbone is divided into a self-decoder that combines full attention and sliding-window attention (SWA), and a cross-decoder that combines full attention and sparse attention.
At the outer level, \textbf{KV Bridging} connects full-attention layers across the two decoders.
Each cross-decoder full-attention layer applies its own K/V projections to the input hidden states of a corresponding self-decoder full-attention layer.
At the inner level, \textbf{KV Reuse} retains HySparse's sharing of KV caches and selection indices within each hybrid block.

HySparse2 also makes two modifications that improve accuracy and efficiency.
First, token-level selection replaces block-level selection, allocating the sparse attention budget more precisely to relevant tokens.
Second, a forced window of recent tokens replaces the separate SWA branch in sparse layers, allowing local and global tokens to use the same shared KV cache.
All cross-decoder KV caches can then be built from self-decoder hidden states.
Prefill can therefore exit after the self-decoder.

We compare HySparse2 with HySparse and Hybrid SWA on 80B-A3B MoE models trained with the same data and schedules. After pretraining, HySparse2 retains broadly comparable general capabilities and improves long-context retrieval. After light post-training, it improves mean MRCR-v2 and RULER-v2 scores over HySparse by 11.30 and 19.81 percentage points, respectively, and achieves lower AgentPPL and LongPPL than both baselines at all evaluated lengths up to 256k. At 1M tokens, our analysis shows 2.92× and 5.02× reductions in prefill FLOPs relative to HySparse and Hybrid SWA, respectively, alongside a smaller KV cache. Ablations show that token-level selection improves retrieval at the same attention budget and that KV Bridging preserves broadly comparable quality.

\section{Background and Motivation}

\subsection{Efficient Prefill}

Long-horizon and multi-turn agents make inference increasingly input-dominated.
Tool responses can add substantially more tokens than the actions that produced them.
Reducing attention cost alone does not eliminate the computation needed to propagate these tokens through the backbone.
Cross-layer KV-cache sharing can reduce both KV storage and prefill computation.
YOCO uses a self-decoder to construct a global KV cache shared across cross-decoder layers, allowing prefill cache construction to exit after the self-decoder~\citep{yoco}.
Gemma 3n adopts related cross-layer KV sharing to improve prefill efficiency~\citep{gemma3n}.
HySparse applies cross-layer sharing within its hybrid sparse-attention blocks: each full-attention layer shares its global KV cache and top-$k$ block indices with the following sparse-attention layers, though prefill still executes all layers~\citep{hysparse}.
HySparse2 further incorporates a YOCO-style structure into HySparse, combining reduced attention computation and KV storage with a shorter prefill path.

\subsection{KV Cache Compression}

Agentic workloads require more compact KV caches.
KV-cache compression spans four dimensions: head, sequence, layer, and precision.
Head-level methods share KV heads through GQA/MQA~\citep{gqa,mqa} or compress KV representations into latent states, as in MLA~\citep{deepseekv2}.
Sequence-level methods compress multiple tokens into fewer cache entries~\citep{deepseekv4}.
Layer-level methods share KV caches across layers~\citep{cla,yoco}.
Precision-level methods reduce the numerical precision of cached keys and values~\citep{liu2024kivi,hooper2024kvquant}.
Prior work focuses on intra-layer compression along the head, sequence, and precision axes.
HySparse2 instead pushes cross-layer sharing through KV Bridging and KV Reuse.

\subsection{Sparse Attention Granularity}

Sparse attention reduces long-context attention costs by restricting each query to a subset of KV entries~\citep{sparsetransformer}.
Its selection granularity creates an inherent trade-off between modeling accuracy and inference efficiency.
HySparse adopted block-level sparsity as a practical compromise, since evaluations showed no substantial accuracy disadvantage and the regular block structure enabled efficient kernels.
Modern agentic workloads, however, involve long-horizon multi-turn trajectories that require precise retrieval from long historical contexts.
Under these patterns, block-level selection exhibits a pronounced accuracy disadvantage, while token-level selection retrieves relevant evidence more faithfully. 
Recent advances in sparse kernels also make token-level implementations practical~\citep{tilelang}.
HySparse2 therefore adopts token-level sparsity for more precise context selection.

\section{HySparse2}

\subsection{Overview}

\begin{figure}[!t]
    \centering
    \includegraphics[width=\linewidth]{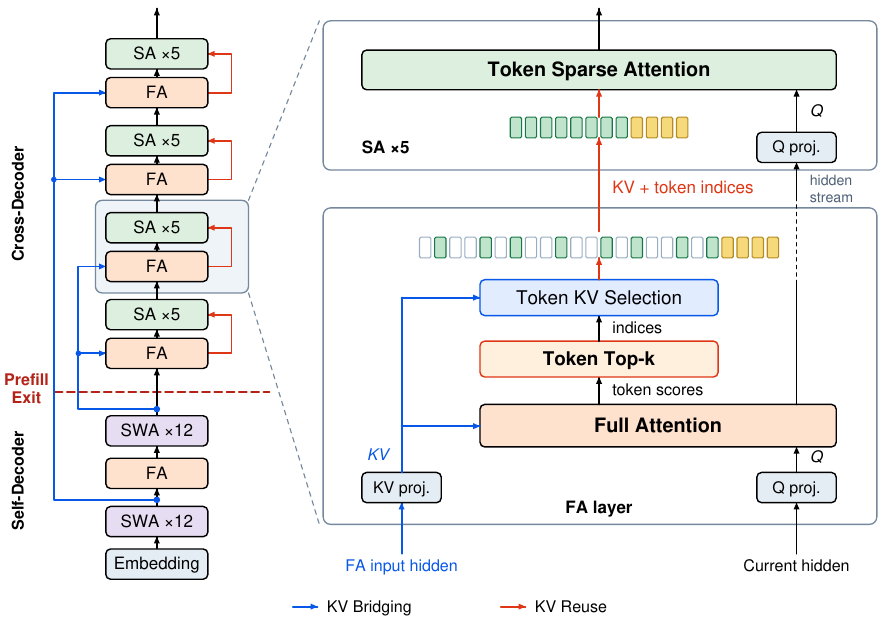}
    \caption{Two-level KV sharing in HySparse2. FA, SWA, and SA denote full attention, sliding-window attention, and sparse attention, respectively. Left: the overall architecture with KV Bridging between the self-decoder and cross-decoder. Prefill cache construction can exit after the self-decoder. Right: a HySparse2 block with token-level sparse attention and a forced local window. Through KV Reuse, SA layers reuse the block’s FA KV cache and selection indices.}
    \label{fig:hysparse2-overview}
\end{figure}

Figure~\ref{fig:hysparse2-overview} illustrates the HySparse2 architecture.
Following YOCO~\citep{yoco}, the backbone is divided into a self-decoder and a cross-decoder.
Both decoders use hybrid attention.
The self-decoder hybridizes full attention with sliding-window attention for local modeling.
The cross-decoder hybridizes full attention with sparse attention for global retrieval.
HySparse2 uses two-level KV sharing.
At the outer level, \textbf{KV Bridging} constructs each cross-decoder full-attention layer's KV cache from corresponding self-decoder hidden states through layer-specific projections.
At the inner level, \textbf{KV Reuse} lets sparse layers reuse the full-attention layer's KV cache and selection indices within each hybrid block.

\subsection{Full-Attention-Only KV Bridging}
\label{sec:kv-bridging}

KV Bridging operates only between full-attention layers in the self-decoder and cross-decoder.
Consider a pair of full-attention layers $(i,j)$, with layer $i$ in the self-decoder and layer $j$ in the cross-decoder.
Let $\mathbf{H}^{\mathrm{self}}_i$ and $\mathbf{H}^{\mathrm{cross}}_j$ denote the corresponding input hidden states.
Cross-decoder layer $j$ computes its keys, values, and queries as
\begin{equation}
    \mathbf{K}^{\mathrm{cross}}_j = \operatorname{Proj}^{K}_{j}\!\left(\mathbf{H}^{\mathrm{self}}_i\right), \quad
    \mathbf{V}^{\mathrm{cross}}_j = \operatorname{Proj}^{V}_{j}\!\left(\mathbf{H}^{\mathrm{self}}_i\right), \quad
    \mathbf{Q}^{\mathrm{cross}}_j = \operatorname{Proj}^{Q}_{j}\!\left(\mathbf{H}^{\mathrm{cross}}_j\right).
\end{equation}
Each cross-decoder full-attention layer has its own K/V projections, so layers that share a hidden-state source still construct distinct KV caches.
Each layer also retains an independent Q projection applied to its own current hidden states.
KV Bridging supports different hybrid attention ratios in the two decoders, allowing one self-decoder full-attention layer to supply multiple cross-decoder full-attention layers, as illustrated in Figure~\ref{fig:hysparse2-overview}.

\subsection{Cross-Layer KV Reuse}
\label{sec:kv-reuse}

HySparse2 makes two key refinements to HySparse: adopting token-level sparse selection and removing the separate SWA branch.

\paragraph{Token-Level Sparse Selection.}

In HySparse, pretraining evaluations showed no substantial accuracy disadvantage for block-level selection.
However, block-level sparsity shows a pronounced accuracy disadvantage on multi-turn, long-context agentic tasks.
HySparse2 therefore applies top-$k$ selection to individual tokens in full-attention layers, and the following sparse layers reuse the KV entries for the selected tokens.
Section~\ref{sec:token-block-ablation} evaluates selection granularity.

\paragraph{Removing the Separate SWA Branch.}

HySparse2 removes the separate SWA branch used in HySparse and supports local modeling by forcing a recent window into the sparse selection.
As illustrated in Figure~\ref{fig:hysparse2-overview}, a local window of the most recent tokens is always selected, followed by the highest-scoring tokens outside the window.
Both selected sets are read from the full-attention KV cache and reused by the following sparse-attention layers.

This design also enables a complete early exit after the self-decoder during prefill. A separate SWA branch in the cross-decoder requires building a suffix of KV cache from projections of its own hidden states, which further depends on an expanding set of hidden states in earlier layers and grows linearly with depth. This cascading SWA dependency therefore requires the cross-decoder to process a token suffix far longer than the window size. The required states must either be computed during prefill, preventing the cross-decoder from being skipped entirely, or approximated by methods such as bounded replay~\citep{deepseekv41flash}. HySparse2 removes this dependency by construction and requires no cross-decoder computation during prefill. Removing the branch also eliminates substantial parameter overhead from the separate projections in our gated baseline.

\subsection{The Role of Full Attention}
Both HySparse2 and HySparse retain a small number of full-attention layers, as we believe that full attention remains important for model quality. 
These full-attention layers also serve as indexers, using exact attention scores to provide oracle token selection for subsequent sparse layers.
This design supports native end-to-end training without a separate indexer or an auxiliary distillation objective to train one.
Full attention nevertheless remains computationally expensive, so we keep the proportion of full-attention layers small. In HySparse2, prefill KV-cache construction requires only one full-attention layer. Following recent sparsification approaches~\citetext{\citealp{seerattn_v1}; \citealp{dsa}}, we could further approximate the retained full-attention layers with a lightweight indexer plus sparse attention in post-training.
Future architectures could also allocate less computation to full attention and more to sparse attention, for example by reducing the number of query heads in full-attention layers while increasing it in sparse-attention layers.

\subsection{Inference Architecture}
\label{sec:prefill-decode}

\paragraph{Prefill–Decode Disaggregation.}

Under prefill--decode disaggregation, the HySparse2 prefill node hosts only the self-decoder and the KV Bridging projections.
For the 49-layer model shown in Figure~\ref{fig:hysparse2-overview}, this requires deploying only the first 25 layers, cutting the memory requirement of the prefill node by nearly half.
With a high SWA-to-full-attention ratio, the self-decoder performs full attention in only one layer during the prefill stage.
We also compute the cross-decoder full-attention KV caches on the prefill node and transfer them to the decode nodes, because the projected KV caches are smaller than the source hidden states.

\paragraph{Speculative Decoding.}

During pretraining, HySparse2 uses a single MTP layer conditioned on the hidden states at the self-decoder/cross-decoder boundary to aid convergence.
During post-training, this MTP layer could be replaced with a larger DFlash-style drafter conditioned on the same hidden states~\citep{chen2026dflash}.
Their early availability could support asynchronous drafting and verification~\citep{zhang2025swiftspec}.
Training the larger drafter and coordinating asynchronous execution with verification feedback remain future work.

\section{Evaluation}

We compare HySparse2 with HySparse~\citep{hysparse} and Hybrid SWA used in MiMo-V2 Series~\citep{xiao2026mimo}.
The overall comparison covers model performance, prefill FLOPs, and KV-cache storage.
We then conduct ablation studies on sparse selection granularity, the forced local window, and KV Bridging. 
Token-level selection targets stronger long-context accuracy in agentic scenarios without compromising efficiency, whereas the forced local window and KV Bridging target lower prefill costs while retaining model quality.

\subsection{Experimental Setup}
\label{sec:eval-setup}

\paragraph{Model configuration.}
Unless otherwise specified, experiments use 80B-A3B MoE models.
Each model has 49 Transformer layers with a hidden size of 2,048 and uses a simplified mHC variant~\citep{zhu2025hyperconnections,xie2026mhc} with the residual mixing matrix fixed to the identity. 

The models only differ in their attention designs, as summarized in Table~\ref{tab:eval-model-config}.
Full attention in every layer is the strongest reference under our evaluation. When choosing how many full-attention layers a hybrid model uses, we avoid a substantial accuracy gap from this baseline. Hybrid SWA retains nine full-attention layers, as using fewer significantly degrades accuracy. HySparse and HySparse2 each use only five full-attention layers while remaining comparable to the full-attention baseline.
Hybrid SWA and HySparse use GQA~\citep{gqa}, whereas HySparse2 uses MQA~\citep{mqa}, which yields a smaller KV cache and better token sparse attention kernel efficiency.

\begin{table}[H]
    \centering
    \small
    \setlength{\tabcolsep}{10pt}
    \begin{tabular}{lccc}
        \toprule
        \textbf{Model} & \textbf{\#Full} & \makecell{\textbf{Heads}\\\textbf{(Q/KV)}} & \makecell{\textbf{Head dim.}\\\textbf{(QK/V)}} \\
        \midrule
        Hybrid SWA & 9 & 64/4 & 192/128 \\
        HySparse & 5 & 64/4 & 192/128 \\
        HySparse2 & 5 & 64/1 & 256/256 \\
        \bottomrule
    \end{tabular}
    \caption{Configurations of the three attention designs.}
    \label{tab:eval-model-config}
\end{table}

In HySparse2, SWA layers in the self-decoder use partial rotary positional embeddings (RoPE)~\citep{rope} with 64 rotary dimensions and a base of 10,000, while full and sparse attention use NoPE.
HySparse2 therefore requires no RoPE adjustment during long-context extension.
Across the three configurations, all sparse and SWA layers use sigmoid output gates~\citep{qiu2025gated} and learnable per-head sink biases~\citep{gpt-oss}.
HySparse2 uses token-level selection with 128 forced local tokens and 1,024 global tokens.
HySparse selects 1,024 global tokens in 64-token blocks and combines sparse attention with a separate 128-token SWA branch through gated fusion.
Hybrid SWA uses a sliding window of 128 tokens.

\paragraph{Training.}
We pretrain the 80B-A3B models on approximately 500B tokens at a context length of 32k.
A light post-training stage adds approximately 100B tokens, introduces agentic data into the training mixture, and extends the context length to 256k.
The three models share the same data mixture and training schedule within each stage.
We use the Muon optimizer~\citep{jordan2024muon} with a WSD schedule and peak learning rates of $10^{-3}$ for pretraining and $5\times10^{-5}$ for post-training.
We report both pretraining and post-training results for overall comparisons, and only pretraining results for ablations.

\paragraph{Evaluation.}

The pretraining evaluation covers knowledge (MMLU, MMLU-Redux, MMLU-Pro, C-Eval, CMMLU, TriviaQA)~\citep{mmlu,gema2024we,wang2024mmlu,huang2023c,li2023cmmlu,joshi2017triviaqa}, reasoning (BBH, MATH, DROP, GSM8K, ARC-C, HellaSwag, WinoGrande)~\citep{suzgun2022challenging,hendrycks2021measuring,dua2019drop,gsm8k,arcc,zellers2019hellaswag,sakaguchi2021winogrande}, code (HumanEval+, MBPP+, Repo Code PPL)~\citep{humaneval,mbpp,liu2023your}, and long-context tasks (RULER, NoLiMa)~\citep{ruler,modarressi2025nolima}, as listed in Table~\ref{tab:v2-v1-pretrain}.
Repo Code PPL is an internal benchmark that reports byte perplexity over long code repositories with cross-file dependencies, and therefore also probes long-context modeling ability.

After light post-training, we focus on multi-turn retrieval and long-context likelihood on agent trajectories.
AgentPPL is an internal benchmark built from multi-turn agent trajectories that measures byte perplexity on the reasoning, tool-call, and response segments of 1,000 trajectories.
LongPPL~\citep{fang2024longppl} measures perplexity on selected tokens that depend on long-range context, using 349 examples from agent trajectories and long-context tasks.
MRCR-v2~\citep{vodrahalli2024michelangelo} measures multi-round retrieval with two, four, or eight needles.
RULER-v2~\citep{ruler_v2_2025} covers 12 retrieval and question-answering subtasks. GraphWalks~\citep{openai2025gpt41} evaluates multi-hop graph traversal.

\subsection{Overall Comparison}
\label{sec:v2-v1}

\paragraph{Pretraining results.}

\begin{table}[t]
    \centering
    \small
    \setlength{\tabcolsep}{10pt}
    \begin{tabular}{lrrr}
        \toprule
        \textbf{Task} & \textbf{Hybrid SWA} & \textbf{HySparse} & \textbf{HySparse2} \\
        \midrule
        \multicolumn{4}{l}{\textit{Knowledge}} \\
        MMLU & 61.48 & \textbf{64.48} & 63.92 \\
        MMLU-Redux & 63.96 & \textbf{68.44} & 67.50 \\
        C-Eval & 65.08 & 67.09 & \textbf{67.90} \\
        CMMLU & 67.41 & \textbf{69.97} & 69.90 \\
        TriviaQA & 60.59 & \textbf{60.67} & 59.67 \\
        \midrule
        \multicolumn{4}{l}{\textit{Reasoning}} \\
        BBH & 60.14 & 61.93 & \textbf{64.29} \\
        MMLU-Pro & 36.12 & 35.74 & \textbf{37.56} \\
        MATH & \textbf{36.18} & 36.14 & 34.32 \\
        DROP & 60.90 & \textbf{63.78} & 58.99 \\
        GSM8K & \textbf{64.44} & 64.14 & 61.94 \\
        ARC-C & 75.51 & \textbf{79.18} & 77.82 \\
        HellaSwag & 79.38 & 79.19 & \textbf{79.54} \\
        WinoGrande & 73.72 & \textbf{74.19} & 71.82 \\
        \midrule
        \multicolumn{4}{l}{\textit{Code}} \\
        HumanEval+ & \textbf{35.98} & 34.76 & 34.15 \\
        MBPP+ & \textbf{55.56} & 52.65 & 52.65 \\
        Repo Code PPL $\downarrow$ & 1.1578 & 1.1588 & \textbf{1.1570} \\
        \midrule
        \multicolumn{4}{l}{\textit{Long context}} \\
        RULER & 88.71 & 84.89 & \textbf{90.77} \\
        NoLiMa & 30.13 & 40.27 & \textbf{49.76} \\
        \bottomrule
    \end{tabular}
    \caption{Pretraining performance of the 80B-A3B models. HySparse2 improves long-context performance while remaining broadly comparable to both baselines on general capabilities.}
    \label{tab:v2-v1-pretrain}
\end{table}

Table~\ref{tab:v2-v1-pretrain} shows that HySparse2 remains broadly comparable to both baselines on general capabilities, while its clearest gains are in long-context modeling. 
HySparse2 achieves the highest RULER and NoLiMa scores, improving over HySparse by 5.88 and 9.49 points, respectively. It also has the lowest Repo Code PPL at 1.1570, a small improvement over 1.1588 for HySparse and 1.1578 for Hybrid SWA. Individual general-purpose results show a mixed picture. HySparse2 is stronger on BBH and MMLU-Pro, while HySparse retains an advantage on DROP.

\paragraph{Agentic and long-context results.}

We next compare the models after the same light post-training stage of approximately 100B tokens.
We evaluate retrieval with MRCR-v2 and RULER-v2 and long-context likelihood with AgentPPL and LongPPL. Retrieval averages weight the reported context lengths equally, and perplexity curves group examples by context length.

\begin{figure}[H]
    \centering
    \includegraphics[width=\linewidth]{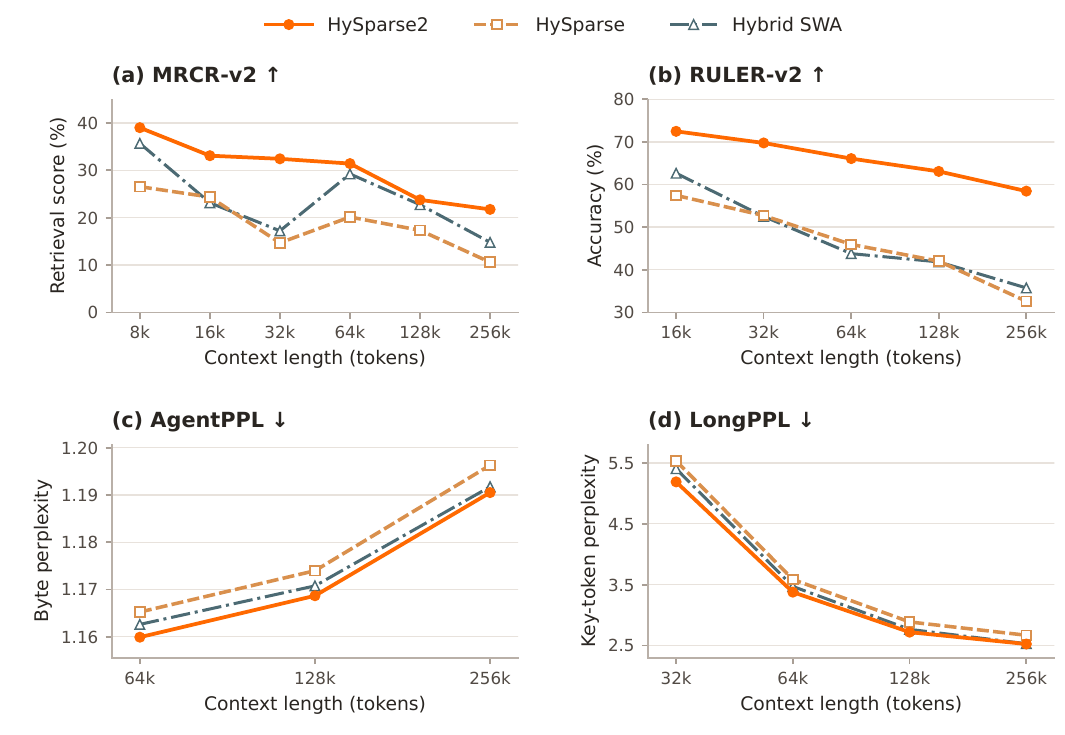}
    \caption{Long-context performance after a light post-training stage. HySparse2 achieves higher retrieval scores and lower perplexity than both baselines across all evaluated context lengths.}
    \label{fig:v2-v1-posttrain}
\end{figure}

Figure~\ref{fig:v2-v1-posttrain} shows that HySparse2 leads both baselines on all four metrics at every evaluated length.
The largest gains are in retrieval.
Relative to HySparse, its mean MRCR-v2 and RULER-v2 scores increase by 11.30 and 19.81 points, while the corresponding gains over Hybrid SWA are 6.44 and 18.65 points.
The advantage remains large at 256k, where HySparse2 reaches 58.45 on RULER-v2, compared with 32.61 for HySparse and 35.74 for Hybrid SWA. 
HySparse2 also achieves lower AgentPPL and LongPPL throughout the evaluated range, showing that its retrieval gains are accompanied by better likelihood on long agent trajectories and context-dependent tokens.
AgentPPL increases with context length, whereas LongPPL decreases.
This contrast reflects opposite effects of additional context: longer contexts expose more relevant information for LongPPL's key tokens, but also introduce more intervening turns and tool outputs that make the relevant evidence harder to retrieve in multi-turn agent trajectories.

\paragraph{Prefill computation and KV-cache storage.}

We compare the prefill FLOPs and KV-cache size of HySparse2, HySparse, and Hybrid SWA, under the 80B-A3B configuration across context lengths from 1k to 1M tokens with FP8 KV-cache storage.

\begin{figure}[h]
    \centering
    \includegraphics[width=\linewidth]{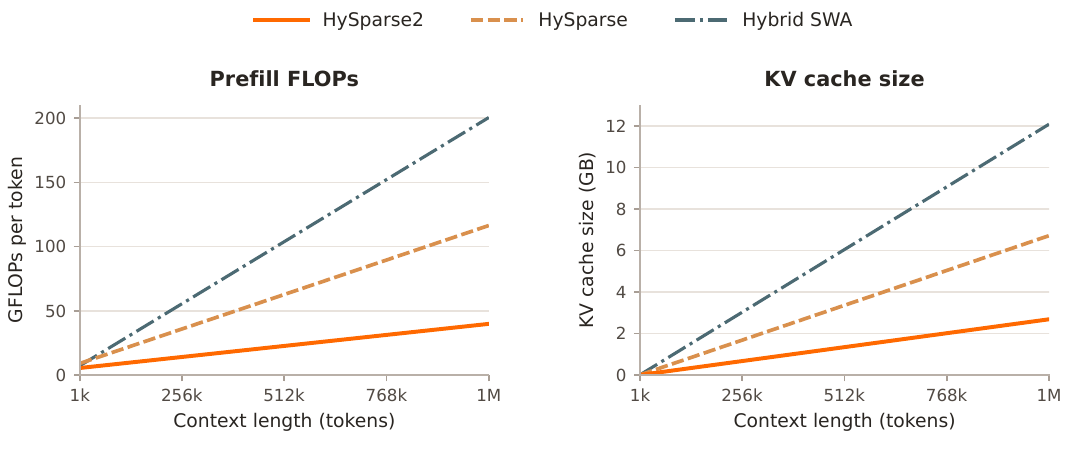}
    \caption{Prefill computation and KV-cache storage for the 80B-A3B models. HySparse2 reduces both costs relative to HySparse and Hybrid SWA.}
    \label{fig:prefill-costs}
\end{figure}

Figure~\ref{fig:prefill-costs} shows that HySparse2 requires less prefill computation and KV-cache storage than HySparse and Hybrid SWA.
At 1M tokens, HySparse2 reduces prefill FLOPs by $2.92\times$ relative to HySparse and $5.02\times$ relative to Hybrid SWA.
The KV cache of HySparse2 occupies only 2.69\,GB, compared with 6.72\,GB for HySparse and 12.09\,GB for Hybrid SWA.

\subsection{Ablation 1: Token-Level versus Block-Level Sparsity}
\label{sec:token-block-ablation}

We compare block-level selection with token-level selection.
Block-level selection uses a block size of 64.
Both variants use the same backbone layout, select 1,024 global tokens, and retain a separate 128-token local window.

\begin{table}[H]
    \centering
    \small
    \setlength{\tabcolsep}{10pt}
    \begin{tabular}{lrr}
        \toprule
        \textbf{Task} & \textbf{Block} & \textbf{Token} \\
        \midrule
        \multicolumn{3}{l}{\textit{Pretraining}} \\
        BBH & \textbf{61.93} & 60.70 \\
        MMLU-Pro & 35.74 & \textbf{36.97} \\
        NoLiMa & \textbf{40.27} & 38.43 \\
        \midrule
        \multicolumn{3}{l}{\textit{Long context ($\leq$32k)}} \\
        RULER-v2 & 49.56 & \textbf{56.13} \\
        MRCR-v2 (2-needle) & 12.94 & \textbf{21.08} \\
        GraphWalks & 29.38 & \textbf{34.92} \\
        \bottomrule
    \end{tabular}
    \caption{Token-level versus block-level sparse selection after pretraining. Token-level selection improves long-context retrieval and graph reasoning under the same attention budget.}
    \label{tab:token-block-ablation}
\end{table}

Table~\ref{tab:token-block-ablation} shows clear gains for token-level selection on long-context retrieval and graph reasoning under the same attention budget.
It improves RULER-v2 by 6.57 points, two-needle MRCR-v2 by 8.14 points, and GraphWalks by 5.55 points.
These gains are present within the 32k training context.

Agent trajectories repeatedly interleave reasoning, tool calls, and returned observations.
The information needed for a response can be spread across several turns, along with role delimiters and other special tokens that identify the relevant boundaries.
Under a fixed attention budget, selecting a block to retain one such token also spends capacity on its neighbors.
Token-level selection can allocate that budget to individual positions across the context.
The retrieval and graph-reasoning gains are consistent with this explanation.

\subsection{Ablation 2: Local Window in Sparse Layers}
\label{sec:swa-ablation}

We compare three ways to handle local context inside sparse layers: a separate gated 128-token SWA branch (\textbf{Gated SWA}), removing the separate SWA branch (\textbf{No SWA}), and forcing the most recent 128 tokens into the sparse selection (\textbf{Forced SWA}).
All three share the same model backbone, KV Bridging, and hybrid SWA self-decoder. They differ only in how the cross-decoder's sparse attention handles local context.
We evaluate pretraining results within their 32k training context.

\begin{table}[H]
    \centering
    \small
    \setlength{\tabcolsep}{10pt}
    \begin{tabular}{lrrr}
        \toprule
        \textbf{Task} & \textbf{Gated SWA} & \textbf{No SWA} & \textbf{Forced SWA} \\
        \midrule
        \multicolumn{4}{l}{\textit{Reasoning}} \\
        BBH & 62.23 & 60.11 & \textbf{62.25} \\
        MMLU-Pro & \textbf{39.15} & 37.19 & 38.12 \\
        MATH & \textbf{35.82} & 33.82 & 33.38 \\
        DROP & \textbf{60.06} & 58.94 & 56.85 \\
        GSM8K & \textbf{64.52} & 60.35 & 59.44 \\
        ARC-C & \textbf{81.06} & 79.01 & 78.41 \\
        HellaSwag & \textbf{79.85} & 78.23 & 78.42 \\
        WinoGrande & 73.09 & \textbf{73.95} & 72.85 \\
        \midrule
        \multicolumn{4}{l}{\textit{Long context ($\leq$32k)}} \\
        NoLiMa & 37.62 & \textbf{38.37} & 36.16 \\
        RULER & 88.19 & 84.55 & \textbf{89.84} \\
        RULER-v2 & 53.66 & 54.62 & \textbf{55.98} \\
        MRCR-v2 & \textbf{27.66} & 20.73 & 22.67 \\
        GraphWalks & 35.39 & 36.48 & \textbf{37.13} \\
        LongPPL $\downarrow$ & \textbf{6.8807} & 7.1307 & 6.9838 \\
        \bottomrule
    \end{tabular}
    \caption{Local-attention ablation in sparse layers. Forced SWA remains competitive with Gated SWA.}
    \label{tab:swa-ablation}
\end{table}

Table~\ref{tab:swa-ablation} shows that local information is necessary for sparse attention. Removing the local branch (No SWA) generally degrades model performance, while Forced SWA remains competitive with Gated SWA on several tasks.
Forced SWA achieves the best RULER, RULER-v2, and GraphWalks scores. Compared with Gated SWA, Forced SWA is 5.08 points lower on GSM8K and 4.99 points lower on MRCR-v2, while LongPPL is 1.50\% higher.
We consider this an acceptable trade-off, as Forced SWA avoids additional projection parameters and requires no local KV cache, thereby making early-exit prefill feasible.

\subsection{Ablation 3: KV Bridging}
\label{sec:kv-mirror-ablation}

\paragraph{With and without KV Bridging.}

KV Bridging is designed primarily to accelerate prefill while preserving model quality. To test whether this design scales, we train 290B-A8B models on approximately 1.8T tokens at a 32k context length, with and without KV Bridging.

\begin{table}[H]
    \centering
    \small
    \setlength{\tabcolsep}{12pt}
    \begin{tabular}{lrr}
        \toprule
        \textbf{Task} & \textbf{w/o bridging} & \textbf{w/ bridging} \\
        \midrule
        MMLU & 72.68 & \textbf{72.80} \\
        TriviaQA & 73.32 & \textbf{74.10} \\
        BBH & \textbf{70.65} & 69.57 \\
        DROP & \textbf{71.37} & 68.17 \\
        GSM8K & \textbf{77.63} & 76.65 \\
        Repo Code PPL $\downarrow$ & \textbf{1.1351} & 1.1353 \\
        RULER & \textbf{96.32} & 96.01 \\
        LongPPL $\downarrow$ & 3.6053 & \textbf{3.4202} \\
        \bottomrule
    \end{tabular}
    \caption{KV Bridging ablation at the 290B-A8B scale. Models with and without KV Bridging achieve broadly comparable quality.}
    \label{tab:kv-mirror-results}
\end{table}

Table~\ref{tab:kv-mirror-results} shows that KV Bridging preserves quality comparable to the baseline without it. MMLU and TriviaQA improve slightly, RULER changes by only 0.31 points, and Repo Code PPL is nearly unchanged. LongPPL improves from 3.6053 to 3.4202. BBH and GSM8K decrease by about one point, and DROP falls from 71.37 to 68.17.

\paragraph{KV Bridging versus KV Mirror.}
For the connection scheme, we compare KV Bridging with KV Mirror~\citep{liu2026hidden} on 80B-A3B models that share the same hybrid backbone and pretraining recipe.
KV Mirror uses U-shaped connections that pair early and late layers in reverse order, $1\!\rightarrow\!n$, $2\!\rightarrow\!n-1$, and so on, whereas KV Bridging connects only the full-attention layers in the self-decoder and the cross-decoder.
Both styles form the target KV representations by re-projecting source hidden states.

\begin{figure}[H]
    \centering
    \includegraphics[width=0.8\linewidth]{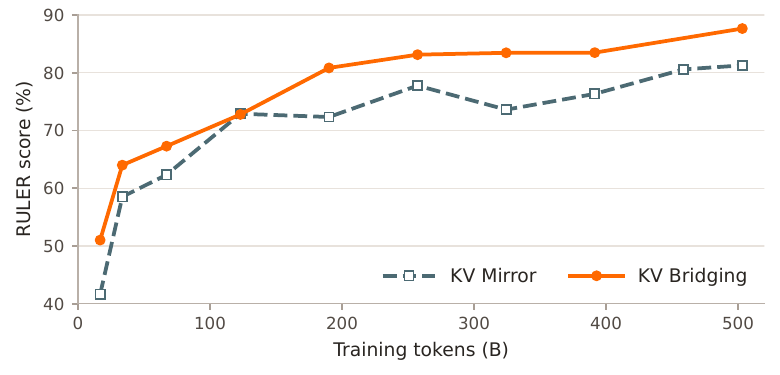}
    \caption{KV Bridging versus KV Mirror during pretraining. KV Bridging achieves higher RULER scores.}
    \label{fig:kv-mirror-connections}
\end{figure}

Figure~\ref{fig:kv-mirror-connections} shows that KV Bridging is stronger for most of training and finishes at 87.65, compared with 81.28 for KV Mirror.
We hypothesize that full-layer states are better sources because full attention backpropagates through scores across the entire visible context, while SWA restricts these direct connections to a local window.
This denser supervision produces representations that retain more global information for projection.

\section{Conclusion}

We present HySparse2, a hybrid sparse attention architecture designed for long-horizon, multi-turn agentic workloads. HySparse2 is a more efficient successor to HySparse and Hybrid SWA, with better long-context and multi-turn retrieval, faster prefill, and lower KV-cache storage.
Its two-level KV sharing enables prefill KV-cache construction by running only half of the model, including just one full-attention layer.
Under prefill–decode disaggregation, the prefill node therefore needs to host only half of the model weights.

\newpage
\bibliography{main}
\newpage
\appendix
\end{document}